\documentclass[letterpaper, 10 pt, conference]{ieeeconf}
\IEEEoverridecommandlockouts
\usepackage{cite}
\usepackage{amsmath,amssymb,amsfonts}
\usepackage{algorithmic}
\usepackage{graphicx}
\usepackage{textcomp}
\usepackage{comment}
\usepackage{xcolor}
\usepackage{censor}

\newif\ifanonymize
\anonymizefalse

\def\BibTeX{{\rm B\kern-.05em{\sc i\kern-.025em b}\kern-.08em
    T\kern-.1667em\lower.7ex\hbox{E}\kern-.125emX}}
\begin{document}

\title{A Measurement-Driven Digital Twin Architecture for Plant-Level Biomass Estimation and Growth Forecasting in Hydroponic Systems}




\ifanonymize
\author{\censor{Morgan Mayborne$^{1}$, Abhisesh Silwal$^{1}$ and George Kantor$^{1}$}}
\else
\author{Morgan Mayborne$^{1}$, Abhisesh Silwal$^{1}$ and George Kantor$^{1}$%
\thanks{This research was funded by the USDA-NIFA Climate-Smart Agriculture for Future Farms (CAFF) program, through award 2023-68017-39402.}%
\thanks{$^{1}$All authors are affiliated with The Robotics Institute, Carnegie Mellon University, Pittsburgh, 15213, USA. Emails: {\tt\small \{mmayborn, asilwal, gkantor\}@andrew.cmu.edu}}%
}

\fi


\maketitle

\begin{abstract}
Alternatives to soil-based horticulture, such as hydroponics, have been developed to respond to food distribution concerns for dense urban centers. A new system was developed to track an individual lettuce plant's growth in a hydroponic environment, utilizing streams of measured information and available models to continuously update the growth trajectory estimates for a plant. These "digital twin" models were integrated into an operating hydroponic greenhouse, with custom horticultural and sensor hardware to grow and measure relevant information. To aid in updating model parameters, plant yield was continuously measured with a custom neural network, using RGB-D images of the plants as an input. The network, trained on a collected dataset of 1300 images, was able to estimate mass within 1.5 g of the ground-truth value. After integration into the custom system, digital twin growth projections could approximate future yield between one and four days in the future, maintaining around a 2 g forecasting error.
\end{abstract}



\section{Introduction}

By enabling precise regulation of environmental variables, hydroponic systems offer advantages relative to conventional soil-based horticulture, including year-round production, vertical stacking for improved space utilization, and more efficient water use \cite{small-medium-hydro}.

Despite these advantages, accurately modeling and forecasting plant growth within hydroponic systems remains a significant challenge. Even under tightly controlled environmental conditions, biological variability across individual plants, nonlinear growth dynamics, and incomplete knowledge of system parameters introduce substantial uncertainty into yield predictions. Prior research has explored growth modeling and optimization in soil-based greenhouse systems \cite{vanthoor2011model}; however, comparatively limited work has investigated adaptive, plant-specific modeling frameworks for hydroponic systems, particularly in cost-sensitive settings.

Digital twins offer a promising framework to address this gap. A digital twin continuously integrates real-time measurements to update internal model states and uncertain parameters, enabling improved prediction accuracy as new data becomes available \cite{digital_twin}. In an agricultural context, the integration of vision-based biomass estimation with dynamic, plant-specific growth models for hydroponic cultivation has not been extensively explored. Such an approach could enable non-destructive growth tracking, adaptive parameter updating, and improved short-term forecasting without requiring complete system identification at initialization.

To address this need, this work presents a plant-specific digital twin framework integrated within an autonomous hydroponic cultivation system for lettuce growth. The proposed system combines environmental sensing, RGB-D imaging, and dynamic growth modeling within a custom greenhouse infrastructure. A convolutional neural network trained on labeled RGB-D images provides continuous, non-destructive biomass estimation, which is assimilated into a dynamic growth model through online parameter updating. This architecture enables real-time state estimation and short-term yield forecasting for individual plants.

The primary contributions of this paper are as follows:
\begin{itemize}
    \item A novel plant-level digital twin architecture for hydroponic growth modeling with continuous data assimilation.
    \item Integration of vision-based biomass estimation with adaptive growth prediction.
    \item Release of a multi-modal dataset comprising environmental measurements, RGB-D imagery, and ground-truth biomass for 125 individual lettuce plants to support future research in plant-level predictive modeling.
\end{itemize}

\section{Background} \label{Background}

A digital twin is a simulation-based representation of a physical system, defined by a mathematical model that continuously updates using real-time data from a dynamic process. Often, these dynamic systems have inherent complexities or unknowns, embedded either in the dynamics of the physical system or in the observation of the physical system. At their core, digital twins require three elements: a physical system, a virtual model, and an update procedure \cite{digital_twin}. The core purpose of digital twins is to build confidence in a value for an otherwise unknown system parameter by using the measured data as feedback for continuous parameter optimization. The "update procedure" is the chosen feedback method for parameter updates, aligning model behavior with real-world behavior. 

\begin{figure}[hbt!]
  \centering
  \includegraphics[width=0.4\textwidth]{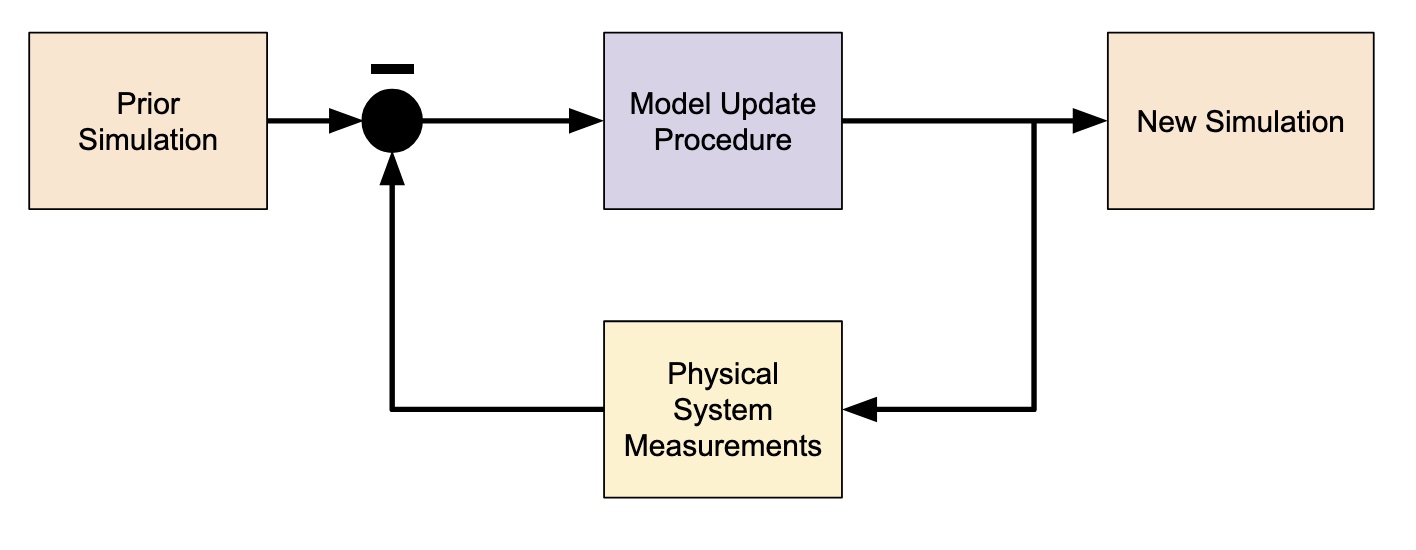}
  \caption{The feedback loop of digital twins, where initial system parameter estimates are updated through an update procedure. Valid update procedures increase confidence in system parameter estimates over time, which is why the loop is pictured as negative feedback.}
  \label{fig:dt-def}
\end{figure}

The modeling component of a digital twin can generally be categorized into three approaches. A white-box approach relies on a fully defined mathematical model of the physical system, with parameters updated deterministically to reduce prediction error. A grey-box approach combines structured physical modeling with probabilistic parameter updates, refining parameter distributions as new measurements become available \cite{dt_probab}. In contrast, a black-box approach uses data-driven models, such as neural networks, to capture complex system dynamics without explicit physical assumptions.

Extending this definition to agriculture introduces two domain-specific challenges: (1) developing growth models that capture complex crop dynamics \cite{crop_modeling}, and (2) autonomously measuring relevant plant and environmental variables \cite{autonomous_measure}. Crop growth is inherently influenced by interacting environmental and physiological factors, requiring carefully structured models. At the same time, reliable measurement of model inputs and outputs is essential. Although environmental variables such as temperature and carbon dioxide are relatively straightforward to monitor, yield estimation often requires more sophisticated mass or volumetric sensing. Effective agricultural digital twins depend on both robust modeling and reliable, non-destructive measurement. The remainder of this section describes how these challenges are addressed in the context of hydroponic lettuce growth.

\subsection{Growth Modeling}

Numerous mathematical models of plant growth can support white- or grey-box digital twin approaches. At the crop scale, models such as CERES \cite{ceres} simulate aggregate field-level growth. Other models operate at the individual plant level, tracking growth indices over time. A notable example is the TOMGRO model, which describes tomato growth as a function of light, carbon dioxide, and temperature \cite{tomgro}.

For lettuce growth modeling, the NiCoLet B3 model adopts a first-order differential equation framework inspired by TOMGRO \cite{nicolet_b3}. The model describes lettuce growth through a carbon balance formulation driven by environmental inputs such as light, temperature, and carbon dioxide, with extensions to account for nitrate-limited conditions.

\subsection{Biomass Evaluation}

Most plant growth models rely on biomass as a primary indicator of development, so a digital twin framework requires a continuous stream of non-destructive biomass measurements to enable feedback and model updating. However, conventional biomass measurement methods are typically destructive, creating a significant challenge for real-time implementation. Developing a reliable, non-destructive biomass estimation pipeline was therefore a central focus of this work. 

This research work investigated phenotyping methods for non-destructive measurement. In particular, image-based analysis techniques enable the measurement of plant attributes such as height, volume, or other structural traits \cite{maize_skeleton}. Among these image-based approaches, Buxbaum et al. developed a convolutional neural network (CNN) architecture to estimate plant biomass from a single overhead RGB-D image captured with a stereo camera, enabling non-destructive and autonomous measurement \cite{buxbaum}. Their approach processed depth and color channels through separate ResNet50 backbones, concatenated the resulting latent representations, and used a multilayer perceptron (MLP) to predict biomass. The method achieved a 7.3\% error on a custom RGB-D dataset.

\section{Methods}

\subsection{Hardware Design}

\begin{figure}[hbt!]
  \centering
  \includegraphics[width=0.3\textwidth]{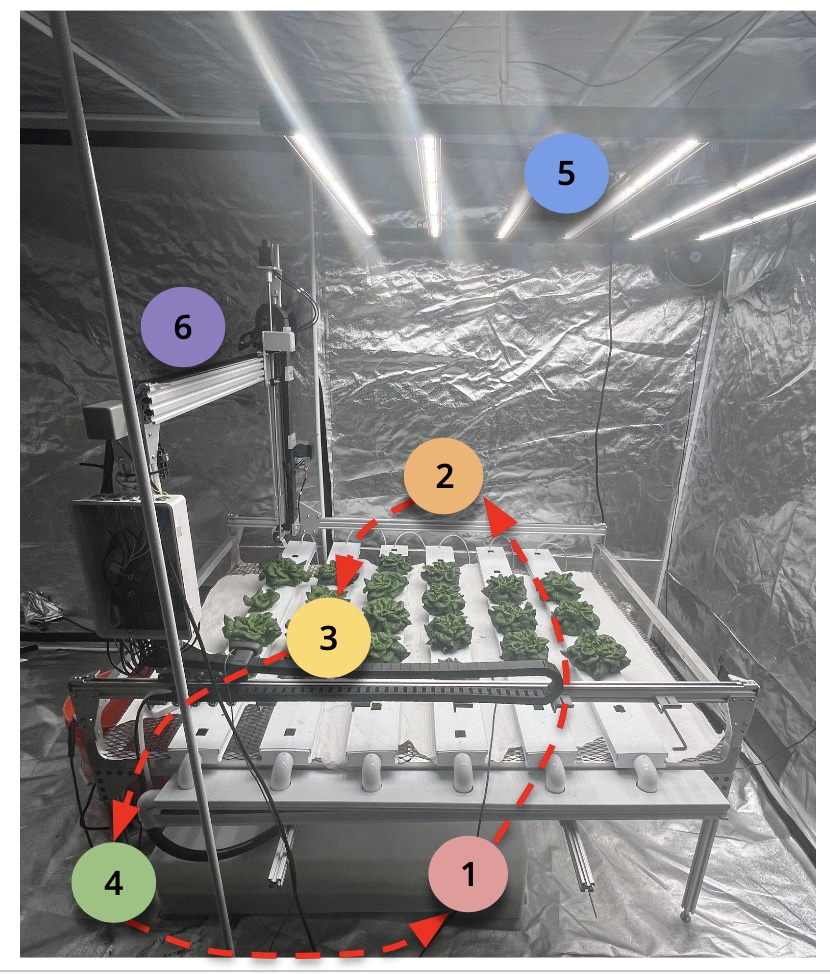}
  \caption{The custom hydroponic growth environment, operating at \censor{Carnegie Mellon University.} (1-4) The NFT nutrient solution circulation flows from the nutrient tank through valved lines into channels, returning to the tank by gravity. (5) A mounted lighting unit is above the planting bed and a (6) FarmBot gantry robot is used for imaging automation.}
  \label{fig:grw-tent}
\end{figure}

The robotic hydroponic cultivation system was deployed inside a sealed grow tent to improve environmental stability. The physical infrastructure consisted of a custom planting bed constructed from extruded aluminum framing. Twenty-four lettuce plants were cultivated using the Nutrient Film Technique (NFT), selected for its hardware simplicity. Rectangular NFT channels were mounted on an aluminum mesh platform above a central nutrient reservoir containing a submersible pump. The pump distributed nutrient solution through valved lines to regulate flow across channels, while gravity returned the solution to the reservoir for recirculation. Nutrient solution electrical conductivity was maintained between 1300--1700~$\mu$S/cm and pH between 5.8 and 6.0. Illumination was provided by a VivoSun LED grow light delivering 200--400~$\mu$mol/m$^2$/s of photosynthetically active radiation (PAR), calibrated using an Apogee MQ-500 quantum sensor.

Environmental data collected for growth modeling included air temperature ($^\circ$C), carbon dioxide concentration (ppm), and photosynthetically active radiation ($\mu$mol/m$^2$/s). Temperature and CO$_2$ were measured using a SAF Aranet4 wireless sensor positioned near canopy level, while light intensity was recorded using an Apogee SQ420 PAR sensor mounted on the planting bed. These variables were selected based on established growth models such as TOMGRO and NiCoLet \cite{tomgro, nicolet_b3}, which identify them as primary drivers of lettuce biomass accumulation.

Autonomous plant monitoring was enabled by mounting a FarmBot Genesis v1.7 gantry robot above the planting bed. An Intel RealSense D405 stereo camera was attached to the robot’s end effector to capture overhead RGB-D images. With the stereo camera, the gantry executed an hourly raster-scan routine to collect consistent plant-level observations. All environmental and imaging data were synchronized and logged locally through custom Python-based control scripts to support the proposed digital twin assimilation.

\subsection{Digital Twin Architecture}

As introduced in Section~\ref{Background}, a digital twin can be broadly defined as a feedback-driven framework in which uncertain model parameters are continuously updated using measured system data. In agricultural applications, this concept depends critically on two components: an appropriate growth model and a reliable measurement pipeline. Accordingly, this section describes the experimental framework used to select a suitable lettuce growth model, implement non-destructive measurement techniques, and define a feedback policy for adaptive parameter updating.

\begin{figure*}[hbt!]
  \centering
  \includegraphics[width=0.6\textwidth]{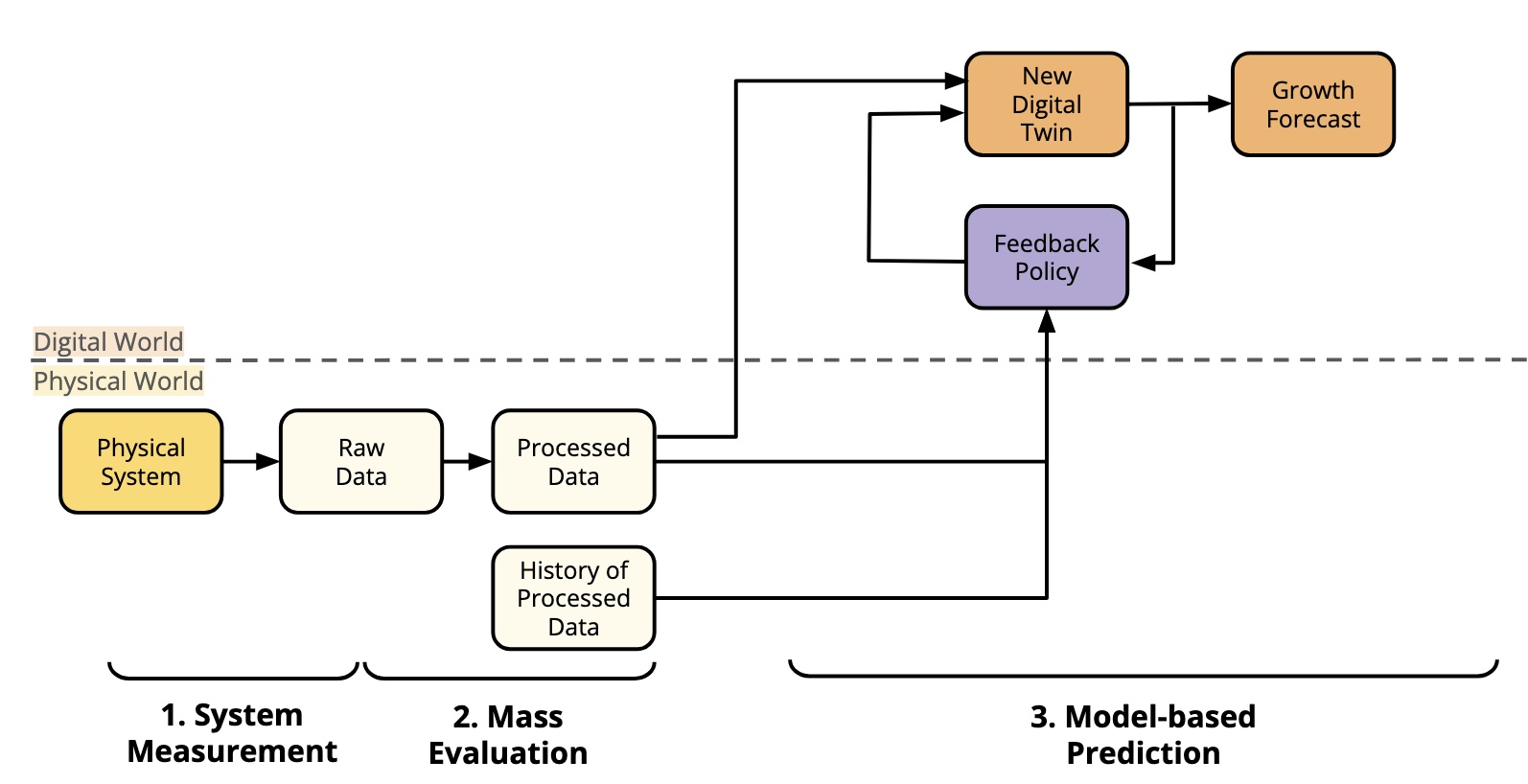}
  \caption{Custom architecture design used for digital twins in this work. Growth environment conditions are measured with raw sensors, with biomass further processed from a raw RGB-D image. These measurements are passed to the digital twin, which can update system parameters to reflect new data.}
  \label{fig:dt-arch}
\end{figure*}

The digital twin architecture implemented in this work is illustrated in Figure~\ref{fig:dt-arch}. The diagram depicts a closed-loop structure linking the physical and digital domains. In the physical world, the hydroponic system generates raw data that are processed into environmental measurements and biomass estimates, which are stored as historical records. In the digital world, these processed measurements are assimilated into a growth model through a feedback policy that updates model parameters. The updated digital twin then produces short-term growth forecasts, completing the recursive loop between sensing, modeling, and prediction.

Additionally, calibration is also necessary to establish strong initial guesses for selected system parameters, which requires tuning on previously collected data. A complete digital twin requires two forms of calibration: (1) biomass measurement models trained with image–mass pairs, and (2) growth models tuned with time-series biomass data.

\subsection{Dataset}

\begin{figure}[hbt!]
  \centering
  \includegraphics[width=0.4\textwidth]{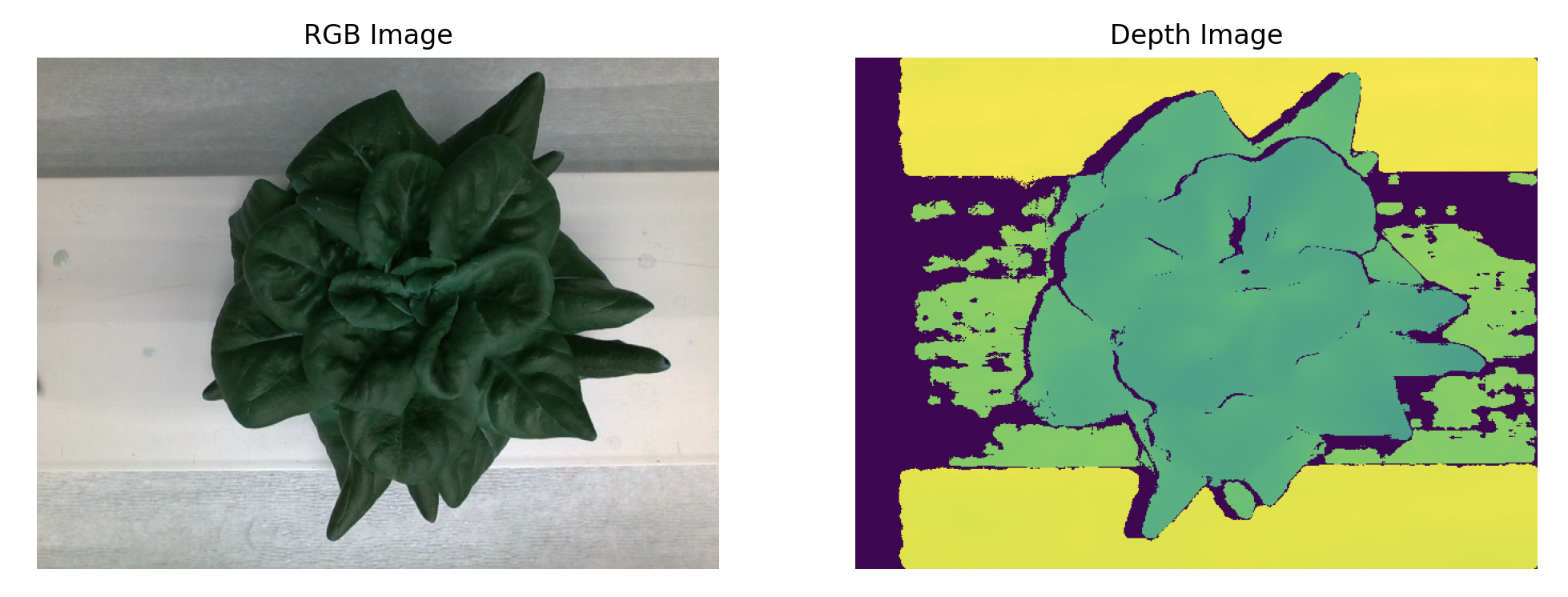}
  \caption{Example of an RGB-D image, with the color and depth channels separated into two images.}
  \label{fig:rgbd}
\end{figure}

To develop the digital twin architecture, a dataset was collected between August and December 2024, tracking the growth of 125 Butterhead lettuce plants over two-week growth cycles. On average, each plant was measured once per day, resulting in longitudinal records of both visual and biomass data. For each measurement, an RGB-D image (resolution: 640 × 480) was captured, and fresh biomass was recorded. Ground-truth mass measurements were obtained by temporarily removing each plant from its NFT channel and weighing it using an off-the-shelf kitchen scale with a resolution of 1 gram. Fresh biomass was defined as plant mass excluding the saturated substrate. Each measurement was logged with a timestamp and stored in a CSV file alongside the corresponding RGB-D image and lighting condition. An example RGB-D image is shown in Figure~\ref{fig:rgbd}.

In total, 1,308 plant measurements were collected, with an average recorded mass of approximately 10g. Each entry in the dataset includes plant identifier, measurement date, fresh biomass, RGB-D image, and associated lighting condition. To improve model robustness and potential generalization, the camera-to-plant distance was varied during data collection. However, a subset of 1,095 images captured at a fixed distance of 175 mm was used for training in this study.

\begin{table}[hbt!]
\centering
 \caption{Dataset Image Distribution}
 \begin{tabular}{|c || c |} 
 \hline
 Attribute & Value \\
 \hline
 Image Resolution & 640 × 480 px \\
 \hline
 Total Number of Images & 1305 \\
 \hline
 Captured at 175mm Distance & 1095 \\
 \hline
 Used for Mass Evaluation Training & 546 \\
 \hline
 Used for Model Calibration & 274 \\
 \hline
 Used for End-to-End Testing & 275 \\
 \hline
 \end{tabular}
 \label{table:dataset-details}
\end{table}

To initialize the digital twin architecture, the dataset was partitioned into separate subsets for module development. 546 samples were reserved to train the biomass evaluation module, while 274 were reserved for calibration of the growth model parameter. The remaining data were held out as an end-to-end test set to evaluate overall system performance on unseen samples. The dataset and associated code base will be made publicly available at: \censor{\textcolor{red}{[https://labs.ri.cmu.edu/aiira/resources/].}}

\subsection{Mass Evaluation}

The proposed digital twin framework requires a reliable biomass estimate for each plant derived from RGB-D imagery. The biomass estimation method must remain robust across varying plant sizes and growth stages, minimizing both average prediction error and variance to ensure stable parameter optimization in subsequent modeling stages. 

\begin{figure}[hbt!]
  \centering
  \includegraphics[width=0.4\textwidth]{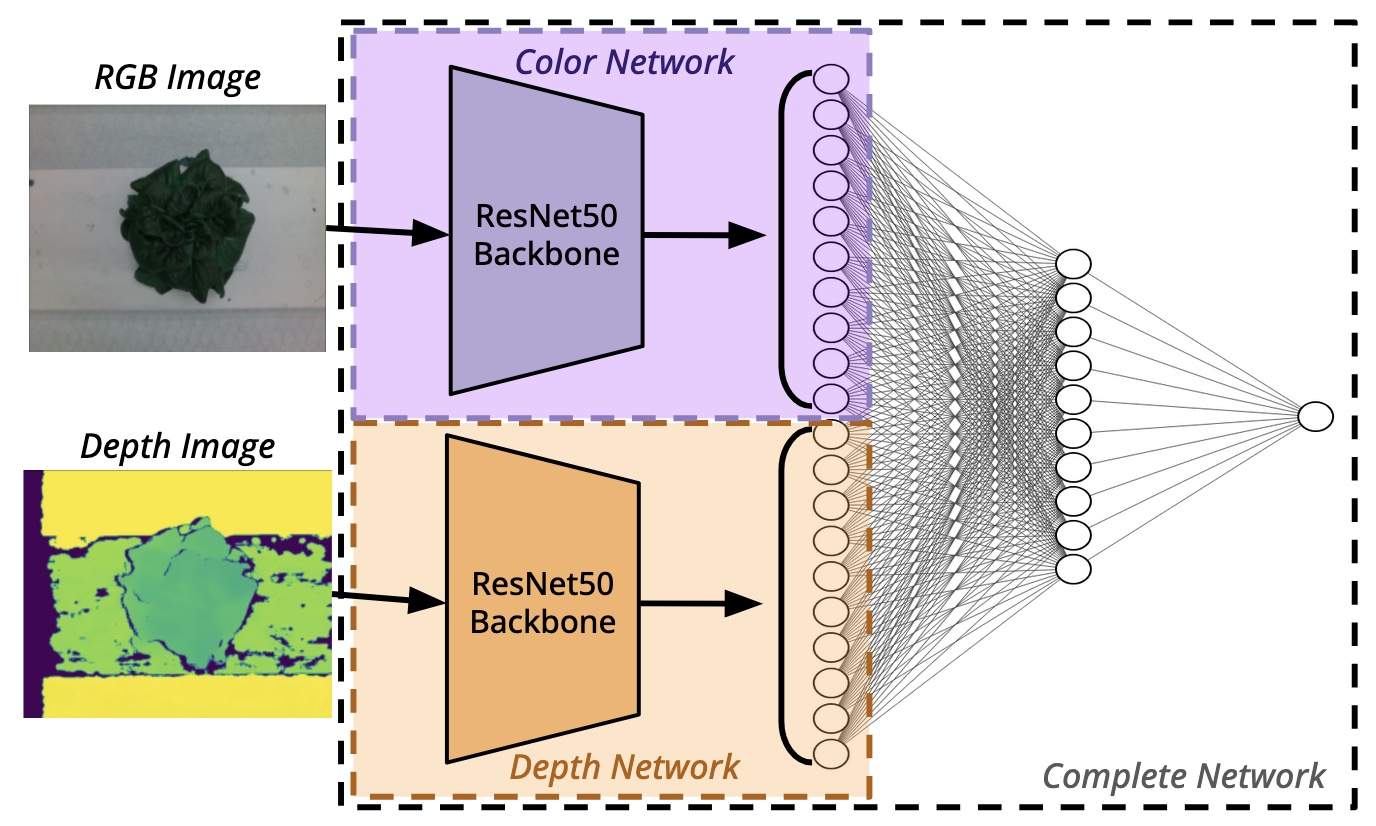}
  \caption{Visualization of the Buxbaum-inspired architecture. RGB and depth channels pass through separate backbones before concatenating latent vectors and connecting to a multilayer perceptron (MLP) to produce a biomass estimate.}
  \label{fig:bux-net}
\end{figure}

Two biomass evaluation strategies were compared. Both methods employed the same neural network architecture (Figure~\ref{fig:bux-net}), inspired by Buxbaum et al., which processes color and depth channels separately before fusing learned representations for final biomass prediction. Specifically, RGB and depth images were passed through independent ResNet50 backbones, and the resulting latent vectors were concatenated and fed into a multilayer perceptron (MLP) to produce a single biomass estimate. \textbf{Method A} trained the full network end-to-end, initializing only the ResNet50 backbones with pretrained weights. \textbf{Method B} adopted a staged training strategy. First, the color and depth subnetworks were trained independently by removing the concatenation layer and connecting each latent representation directly to a single biomass output. This allowed each subnetwork to specialize in extracting relevant features from its respective modality. After this pretraining phase, the complete fused network was initialized using the learned subnetwork weights and then fine-tuned end-to-end in the same manner as Method A. This comparison evaluates whether modality-specific pretraining improves biomass estimation performance relative to direct end-to-end training.

The proposed architectures were implemented in Python using the PyTorch framework. Ground-truth mass–image pairs from the training set were used to train and calibrate the architecture. Model performance was assessed using root mean squared error (RMSE) and the coefficient of determination ($R^2$) between predicted and measured biomass values.

\subsection{Growth Modeling}

Given new measurements, this digital twin architecture should be able to update a model's behavior to better reflect new information. To accomplish this, adjustable model parameters must be defined and tuned to minimize modeling error, which is defined as the residual between the measured and simulated time series. This section will detail an experiment that observes three growth models and their ability to respond to measured data. Experiments with these growth models will be conducted on offline data from the collected dataset, and a proposed formulation for online model updates will also be provided. Performance will be assessed by the average prediction accuracy across different prediction horizons.

\subsubsection{Exponential Model}

A simple exponential model was used to approximate early-stage lettuce growth. Although plant growth ultimately saturates near maturity, this study focused on the first 2–4 weeks of development, during which biomass accumulation can be reasonably approximated by exponential behavior. The model is defined as:

\begin{equation} \label{eq:1}
M_{FM} = Ae^{Bt}
\end{equation}

where \textit{A} represents the initial biomass and \textit{B} denotes the exponential growth rate. Both parameters are treated as adjustable variables within the digital twin and are optimized to minimize modeling error. To estimate parameters, the exponential model is fit over a recent window of measurements using linear least-squares regression in the logarithmic domain. Future biomass values are then predicted by extrapolating the fitted exponential curve. 



\subsubsection{Biological Model}

To represent a white-box modeling approach, the NiCoLet B3 model \cite{nicolet_b3} was evaluated as a biologically grounded growth model for lettuce. This model formulates plant growth as a first-order differential equation system based on carbon balance dynamics. The governing equations are provided in Appendix~\ref{appendA}.

The model tracks carbon in two compartments: carbon stored in the vacuole ($M_{cv}$) and carbon incorporated into plant structure ($M_{cs}$). The rate of change of these states is governed by four physiological processes: photosynthetic assimilation ($F_{cav}$), growth ($F_{cvs}$), maintenance respiration ($F_{cm}$), and growth respiration ($F_{cg}$). Photosynthetic assimilation converts light and carbon dioxide into carbon-based sugars, increasing vacuole carbon. Growth transfers carbon from the vacuole to structural biomass. Maintenance and growth respiration consume available carbon to sustain and expand plant tissues. Environmental control inputs: temperature ($T$), light intensity ($I$), and carbon dioxide concentration ($C$), modulate the intensity of these processes.

Model states, expressed as carbon mass, are converted into fresh and dry biomass outputs ($M_{FM}$ and $M_{DM}$) through transformation equations (Equations~\ref{eq:14}–\ref{eq:15}). Sensitivity analysis by Lopez-Cruz et al. \cite{nicolet_sens} identified three parameters with the greatest influence on model behavior: the maintenance respiration coefficient ($k$), growth rate coefficient ($v$), and leaf area closure parameter ($a$). In this work, these parameters were calibrated to minimize modeling error relative to measured biomass time series.

Parameter optimization was performed using a gradient-free Nelder–Mead method to avoid instability associated with the model’s exponential terms. Parameter values were initialized from the NiCoLet B3 formulation ($k = 4 \times 10^{-7}$, $v = 22.1$, $a = 0.2$). Future biomass predictions were generated by extrapolating the calibrated model forward in time. For online updating, parameter optimization may be performed using batched time series from multiple plants to reduce overfitting, or by applying bounded update steps to ensure stability.

\subsubsection{Neural-Network Model}

A neural network approach was explored to capture system dynamics not fully represented by the biological model. A key challenge in this application is the limited dataset, as plant growth is slow and measurements are collected infrequently. This low-data regime requires models that can learn complex temporal relationships efficiently. A Long Short-Term Memory (LSTM) network was selected due to its ability to model sequential growth patterns and its data-efficient training behavior observed in preliminary experiments compared to multilayer perceptron (MLP) and transformer-based architectures.

The LSTM input consisted of a two-column time-series array containing consecutive daily measurements of lighting and biomass in the first and second columns, respectively. Although additional environmental variables could be incorporated, other conditions were held constant in this study. The input window length was treated as a tunable hyperparameter and was initially set to four consecutive days of measurements.

To evaluate the LSTM-based approach, a four-layer neural network architecture was designed. The model consists of two stacked LSTM layers followed by two MLP layers. The input time-series array is first processed by an LSTM layer with 64 hidden units, followed by a second LSTM layer with 32 hidden units. The resulting 32-dimensional latent representation is then passed through the MLP layers to produce a single scalar output corresponding to the predicted biomass at a specified prediction horizon. The network was trained using mean squared error (MSE) as the loss function.

To generate future biomass forecasts, a new input window of recent measurements is passed through the trained network to produce a predicted mass at the specified prediction horizon. For online adaptation, trained model weights can be stored locally and updated using incremental learning as new plant measurements become available.

\section{Results}

\subsection{Mass Evaluation}

Methods A and B used an architecture inspired by Buxbaum et al., which extracts features from color and depth channels separately before fusing them for biomass prediction (Figure~\ref{fig:bux-net}). Method A trained the full network end-to-end and achieved an average error of 2.03g. Method B pretrained the modality-specific subnetworks before fine-tuning the complete model, reducing the average error to 1.91g.

\renewcommand{\arraystretch}{1.25}
\vspace{-10pt}
\begin{table}[hbt!]
\centering
 \caption{Performance Comparison for Training Strategies}
 \begin{tabular}{|c || c | c |} 
 \hline 
  & RMSE (in grams) & $\mathrm{R}^2$ \\ 
 \hline
 Buxbaum [from scratch] & 2.03 & 0.833 \\
 \hline
 Buxbaum [pre-trained subnetworks] & 1.91 & 0.875 \\
 \hline
 \end{tabular}
 \label{table:method-comp}
\end{table}

Building on Method B, three ablation studies were conducted to further improve biomass estimation performance. 

\subsubsection{Color Segmentation} 
To reduce background noise, HSV-based color segmentation was applied to isolate plant pixels. Non-green pixels were masked, followed by morphological closing, and assigned nominally large depth values. As shown in Table~\ref{table:color-seg-ablation}, segmentation reduced RMSE from 1.91g to 1.56g and improved $R^2$ from 0.875 to 0.915.

The results are shown in Table~\ref{table:color-seg-ablation}, which also highlights that color segmentation improved the mass estimation error, improving the RMSE from 1.91g to 1.56g.

\vspace{-10pt}
\begin{table}[hbt!]
\centering
 \caption{Performance Comparison with Added Color Segmentation}
 \begin{tabular}{|c || c | c |} 
 \hline
  & RMSE (in grams) & $\mathrm{R}^2$\\
 \hline
 Base Performance & 1.91 & 0.875 \\
 \hline
 Color Segmentation & 1.56 & 0.915  \\
 \hline
 \end{tabular}
 \label{table:color-seg-ablation}
\end{table}

\subsubsection{Backbone Substitution} 
To evaluate whether newer CNN backbones improved performance, ResNet50 was replaced with DenseNet121 \cite{densenet} and EfficientNetv2-S \cite{effnet}. Results (Table~\ref{table:backbone-ablation}) showed DenseNet121 achieved the best performance, reducing RMSE from 1.56 g to 1.45 g ($R^2 = 0.929$). DenseNet121 was therefore selected as the final backbone.

\vspace{-10pt}
\begin{table}[hbt!]
\centering
 \caption{Performance Comparison for Backbone Architectures}
 \begin{tabular}{|c || c | c |} 
 \hline
  & RMSE (in grams) & $\mathrm{R}^2$\\
 \hline
 ResNet50 & 1.56 & 0.915  \\
 \hline
 DenseNet121 & 1.45 & 0.929  \\
 \hline
 EfficientNetv2-S & 1.48 & 0.920  \\
 \hline
 \end{tabular}
 \label{table:backbone-ablation}
\end{table}

\subsubsection{Data Augmentations} 
Lastly, geometric (flipping, cropping, rotation) and color-based augmentations (grayscale conversion, HSV jitter) were evaluated (Table~\ref{table:aug-ablation}). Neither augmentation strategy improved performance beyond the 1.45g RMSE baseline. Overall, these ablation studies reduced the biomass estimation error to 1.45g with an $R^2$ of 0.929.

\vspace{10pt}
\begin{table}[hbt!]
\centering
 \caption{Performance Comparison for Image Augmentations}
 \begin{tabular}{|c || c | c |} 
 \hline
  & RMSE (in grams) & $\mathrm{R}^2$\\
 \hline
 Base Performance & 1.45 & 0.929 \\
 \hline
 Geom. Augmentations & 1.71 & 0.900  \\
 \hline
 Color Augmentations & 1.45 & 0.925  \\
 \hline
 \end{tabular}
 \label{table:aug-ablation}
\end{table}

From these ablation studies, the mass evaluation network improved to an average error of 1.45 g and $\text{R}^2$ of 0.929.

\subsection{Growth Modeling}

This section compares the prediction accuracy of the evaluated growth models. Each model was calibrated using measurements from the collected dataset. To simulate real-world deployment, where biomass estimates from the CNN serve as the only available measurements, the models were tested using CNN-predicted masses from the held-out test set. Prediction performance was quantified as the average absolute error between predicted and ground-truth biomass values.

\subsubsection{Exponential Model} 
This approach did not require explicit parameter calibration, as the exponential model was fit to a window of measurements and extrapolated to predict future biomass. Table~\ref{table:method-comp} summarizes the performance of the exponential model across different prediction horizons.

\subsubsection{Biological Model}  
Using the Nelder–Mead optimization procedure described previously, the NiCoLet model parameters were calibrated to the following values:

\begin{itemize}
    \item Maintenance Respiration Coefficient (K) = 4e-7
    \item Growth Rate Coefficient (V) = 22.8
    \item Leaf Area Closure parameter (A) = 0.2
\end{itemize}

Although only a slight adjustment was made to the growth rate coefficient, the optimized parameter set produced a notable improvement in prediction accuracy on the training data. A comparison between the baseline and calibrated models is presented in Table~\ref{table:nicolet-comp}.

\vspace{-10pt}
\begin{table}[hbt!]
\centering
 \caption{NICOLET Prediction Accuracy with Calibrated Parameters}
 \begin{tabular}{|c | c | c | c | c |} 
 \hline 
   & 1-Day & 2-Day & 3-Day & 4-Day \\
   & Horizon & Horizon & Horizon & Horizon \\
 \hline
 Baseline NiCoLet & 2.41g  & 2.50g & 2.42g & 2.75g   \\ 
 \hline
 Calibrated NiCoLet & 1.88g  & 2.05g & 1.89g & 2.22g    \\
 \hline
 \end{tabular}
 
 \label{table:nicolet-comp}
\end{table}

\vspace{10pt}
\subsubsection{Neural-Network Model} 

Using the defined LSTM architecture, networks were trained with varying input window lengths to assess sensitivity to temporal context. Specifically, three models were evaluated to determine whether a 4-day input window led to overfitting or underfitting. Results of this ablation study (Table~\ref{table:window-comp}) indicated that the 4-day window overfit the available data. The optimal performance was achieved with a 2-day input window, suggesting that lettuce growth dynamics may exhibit \textbf{ \textit{second-order Markov behavior}} under the experimental conditions. While further validation is required, the 2-day window configuration was selected for subsequent performance comparisons.

\begin{table}[hbt!]
\centering
 \caption{LSTM Prediction Accuracy for Varying Window Sizes}
 \begin{tabular}{|c | c | c | c | c |} 
 \hline 
   & 1-Day & 2-Day & 3-Day & 4-Day \\
   & Horizon & Horizon & Horizon & Horizon \\
 \hline
 2-Day Window & 1.86g  & 2.27g & 2.19g & 2.21g \\
 \hline
 4-Day Window & 2.10g  & 2.41g & 2.21g & 2.59g   \\
 \hline
 6-Day Window & 2.16g  & 2.44g & 2.84g & 3.57g    \\
 \hline
 \end{tabular}
 \label{table:window-comp}
\end{table}

\begin{figure}[hbt!]
  \centering
  \includegraphics[width=0.35\textwidth]{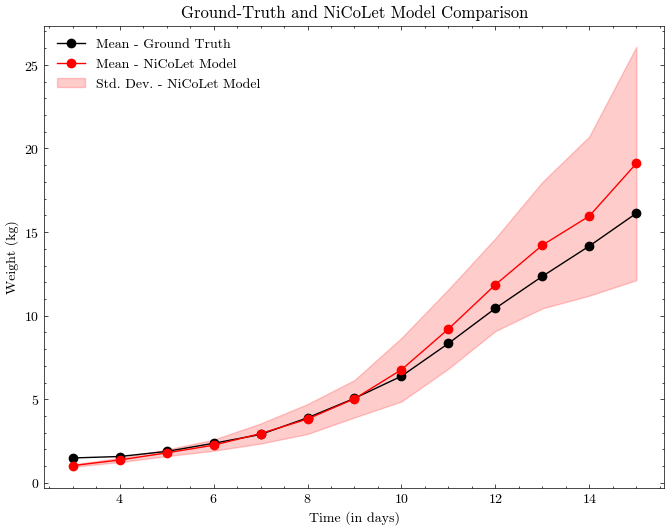}
  \caption{Comparison between measured and NiCoLet-simulated growth, averaged across all samples in the end-to-end testing subset. The shaded region represents the variance in model simulations, primarily attributable to variations in lighting conditions within the greenhouse.}
  \label{fig:sim-comp}
\end{figure}

Table~\ref{table:method-comp} summarizes the prediction performance of the growth models evaluated in this study. Although both the calibrated NiCoLet model and the LSTM network achieved comparable short-term accuracy, the NiCoLet model maintained more consistent performance across increasing prediction horizons. In contrast, the LSTM model exhibited a noticeable decline in accuracy as the forecasting horizon extended. Based on this robustness to longer-term extrapolation, the NiCoLet model was selected as the preferred growth model for the digital twin. Figure~\ref{fig:sim-comp} illustrates the agreement between the calibrated NiCoLet simulations and the aggregate ground-truth growth trajectories.

\vspace{-10pt}
\begin{table}[hbt!]
\centering
 \caption{Prediction Accuracy Between Evaluated Growth Models}
 \begin{tabular}{|c | c | c | c | c |} 
 \hline 
   & 1-Day & 2-Day & 3-Day & 4-Day \\
   & Horizon & Horizon & Horizon & Horizon \\
 \hline
 Exponential & 2.20g  & 2.41g & 2.61g & 4.44g  \\
 \hline
 Calibrated NiCoLet & 1.88g  & 2.05g & 1.89g & 2.22g    \\
 \hline
 LSTM & 1.86g  & 2.27g & 2.19g & 2.21g   \\
 \hline
 \end{tabular}
 
 \label{table:method-comp}
\end{table}

\section{Future Work}

This work developed a digital twin architecture for forecasting lettuce growth in a controlled greenhouse environment, including offline parameter calibration and a formulation for online updates. Future work will focus on implementing and refining real-time parameter adaptation to improve forecasting accuracy as new measurements become available.

A limitation of the current system is the use of proprietary lighting and ventilation hardware, which restricts direct environmental control. Future research could integrate custom hardware to enable closed-loop environmental optimization and evaluate model performance under varying growth conditions. Although demonstrated for hydroponic lettuce using the NiCoLet model, the framework can be extended to other crops by incorporating alternative growth models, such as TOMGRO for tomato plants.

\section{Conclusion}

This work developed a plant-level digital twin architecture for hydroponic lettuce growth forecasting, integrating RGB-D biomass estimation with adaptive growth modeling. Among the evaluated approaches, the calibrated NiCoLet model provided the most robust performance across prediction horizons. These results demonstrate the feasibility of measurement-driven digital twins for improving short-term yield prediction in controlled-environment agriculture.

\ifanonymize
\else
\section*{Acknowledgment}

The authors thank the members of the Kantor Lab at Carnegie Mellon University, as well as our research collaborators at Virginia Tech led by Prof. Song Li, for their valuable feedback and support throughout this research.
\fi

\bibliographystyle{IEEEtran}
\bibliography{references}

\appendix[Growth Model Definition]\label{appendA}

\begin{equation} \label{eq:2}
p(I, C) = \frac{\epsilon \cdot I \cdot \sigma \cdot (C-C^*)}{\epsilon \cdot I + \sigma \cdot (C-C^*)}
\end{equation} 
\begin{equation} \label{eq:3}
f(M_{cs}) = 1 - e^{-a M_{cs}}
\end{equation}
\begin{equation} \label{eq:4}
h_p(M_{cs}, M_{cv}) = \left( 1+ \left( \frac{(1-b_p)\cdot\Pi_v}{\Pi_v \cdot \gamma \cdot C_{cv}(M_{cs}, M_{cv})} \right) ^{s_p}\right)^{-1}
\end{equation}
\begin{equation} \label{eq:5}
F_{cav} = p(I, C) \cdot f(M_{cs}) \cdot h_p(M_{cs}, M_{cv}) 
\end{equation}
\begin{equation} \label{eq:6}
e(T) = K \cdot e^{c (T-T^*)}
\end{equation}
\begin{equation} \label{eq:7}
F_{cm} = e(T) \cdot M_{cs}
\end{equation}
\begin{equation} \label{eq:8}
g(T) = v\cdot e(T)
\end{equation}
\begin{equation} \label{eq:9}
h_g(M_{cs}, M_{cv}) = \left( 1+ \left( \frac{b_g\cdot\Pi_v}{\gamma \cdot C_{cv}(M_{cs}, M_{cv})} \right) ^{s_g}\right)^{-1}
\end{equation}
\begin{equation} \label{eq:10}
F_{cvs} = g(T) \cdot f(M_{cs}) \cdot h_g(M_{cs}, M_{cv})
\end{equation}
\begin{equation} \label{eq:11}
F_{cg} = \theta \cdot F_{cvs}
\end{equation}
\begin{equation} \label{eq:12}
\dot M_{cv} = F_{cav}-h_g(M_{cs}, M_{cv}) \cdot F_{cm}-F_{cg}-F_{cvs}
\end{equation}
\begin{equation} \label{eq:13}
\dot M_{cs} = F_{cvs}-(1-h_g(M_{cs}, M_{cv})) \cdot F_{cm}
\end{equation}
\begin{equation} \label{eq:14}
M_{DM} = \eta_{OMC}\cdot(M_{cv}+M_{cs})+\eta_{MMN} \cdot (\frac{\lambda \cdot \Pi_v}{\beta} \cdot M_{cs}-\frac{\gamma}{\beta} \cdot M_{cv})
\end{equation}
\begin{equation} \label{eq:15}
M_{FM} = 1000 \cdot \gamma \cdot M_{cs} + M_{DM}
\end{equation}

\end{document}